# A Processing Model for Free Word Order Languages

Owen Rambow and Aravind K. Joshi
University of Pennsylvania
rambow,joshi@linc.cis.upenn.edu

# INTRODUCTION

German is a verb-final language. Like many verb-final languages, such as Hindi, Japanese, and Korean, it displays considerable word-order freedom: there is no syntactic constraint on the ordering of the nominal arguments of a verb, as long as the verb remains in final position. This effect is referred to as "scrambling", and is interpreted in transformational frameworks as leftward movement of the arguments. Furthermore, arguments from an embedded clause may move out of their clause; this effect is referred to as "long-distance scrambling". While scrambling has recently received considerable attention in the syntactic literature, the status of long-distance scrambling has only rarely been addressed. The reason for this is the problematic status of the data: not only is long-distance scrambling highly dependent on pragmatic context, it also is strongly subject to degradation due to processing constraints. As in the case of center-embedding, it is not immediately clear whether to assume that observed unacceptability of highly complex sentences is due to grammatical restrictions, or whether we should assume that the competence grammar does not place any restrictions on scrambling (and that, therefore, all such sentences are in fact grammatical), and the unacceptability of some (or most) of the grammatically possible word orders is due to processing limitations. In this paper, we will argue for the second view by presenting a processing model for German.

German is an interesting language to study from the point of view of both competence syntax and performance because it not only allows scrambling, but also topicalization of arguments. Topicalization refers to the movement of a single element into the sentence-initial position in the root clause. Because German is a verb-second language, in every sentence, some element must topicalize. Like scrambling (and like topicalization in English), topicalization in German can create unbounded dependencies. However, the two types of movement differ in terms of their linguistic properties; for example, scrambling can create new anaphoric bindings, while topicalization cannot (Webelhuth, 1989). In addition to the linguistic differences, there is also a processing difference: long-distance topicalization into sentence-initial position appears to be easier to process than long-distance scrambling of the same element over a similar distance. A simple processing account that somehow measures the number of intervening lexical items must fail.

Thus, not only does the German data call for a processing model, but the model must be sensitive to subtle differences in the constructions involved. Our processing model for free-word order languages has two important properties:

- The processing model provides a metric that makes predictions about processing difficulty on an open-ended scale. This property allows us to verify our model with respect to the results from psycholinguistic experiments as well as from native-speaker intuition.

- The processing model is tightly coupled with the competence grammar, in the sense that the grammar directly determines the behavior of the parser. This tight coupling means that if two superficially similar constructions have rather different linguistic analyses, then their processing behavior may well be predicted to be different.

The paper is structured as follows. In the next section, we discuss the relevant issues in German syntax and isolate two phenomena for which we wish to derive a processing model. We then present a grammar formalism (TAG) and an associated automaton (BEPDA). After that, we give linguistic examples and show that the model makes correct predictions with respect to certain cross-linguistic psycholinguistic data. We thereupon discuss the extensions to the basic model that are required to handle long-distance scrambling. We show how the extended model makes plausible predictions for the two phenomena that we identify in the data section.[1]

# GERMAN DATA

German is a verb-final language, but in addition it is verb-second, which means that in a root clause, the finite verb (main verb or auxiliary) moves into the second position in the clause (standardly assumed to be the COMP position).[2] This divides the root clause into two parts: the position in front of the finite verb, the *Vorfeld* or Forefield (VF), and the positions between the finite and non-finite verbs, the *Mittelfeld* or Middlefield (MF). The VF must contain exactly one constituent, which can be any element (an argument or an adjunct, or the non-finite verb). Three types of word-order variation ("movement" in transformational frameworks) are possible: in *extraposition*, embedded clauses appear behind the non-finite verb; *topicalization* fills the VF with an element from the MF; and *scrambling* permutes the elements of the MF. We will discuss them in turn.

---

[1] This work was partially supported by the following grants: ARO DAAL 03-89-C-0031; DARPA N00014-90-J-1863; NSF IRI 90-16592; and Ben Franklin 91S.3078C-1. We would like to thank an anonymous reviewer for very insightful and helpful comments, and Michael Niv for helpful discussions.

[2] Note that in the case of clauses with simple tensed verbs, the final position in the clause for the non-finite verb remains empty.

### Extraposition

While nominal arguments must appear in the MF, clausal arguments may appear behind the verb in clause-final position. (In fact, finite subclauses *must* appear in this position.) An example:

(1) a. ...daß Peter dem Kunden den Kühlschrank zu reparieren zu helfen versucht
    ...that Peter the client (DAT) the refrigerator (ACC) to repair to help tries

    ...that Peter tries to help the client repair the refrigerator

   b. ...daß Peter versucht, dem Kunden zu helfen, den Kühlschrank zu reparieren

   c. ...daß Peter versucht, dem Kunden den Kühlschrank zu reparieren zu helfen

(1a) shows the unextraposed, center-embedded order. (1b) is the fully extraposed order, while (1c) shows that an extraposition of a center-embedded two-clause structure is possible. The fully extraposed version is by far the preferred one in both spoken and written German, especially in situations with more than two clauses.

### Topicalization

We give some examples of topicalization:

(2) a. Der Lehrer hat den Kindern dieses Buch gegeben
    the teacher (NOM) has the children (DAT) this book (ACC) given

    The teacher has given this book to the children

   b. Dieses Buch hat der Lehrer den Kindern gegeben

In (2a), the default word order, the subject is topicalized into the VF. In (2b), the direct object has topicalized into the VF, so that the subject remains in the MF. An adjunct could also occupy the VF, so that all three arguments of *geben* 'to give' would be in the MF. In the case of embedded clauses, the V-2 phenomenon does not occur if there is an overt complementizer, and hence there is no topicalization. The finite verb is in clause-final position. An example:

(3) Ich glaube, daß der Lehrer den Kindern dieses Buch gegeben hat
    I think that the teacher(NOM) the children (DAT) this book (ACC) given has

    I think that the teacher has given this book to the children

If we are only interested in scrambling or extraposition, we will give examples of subordinate clauses so that we do not have to deal with the orthogonal issue of topicalization.

Can elements topicalize out of embedded clauses? In the presence of a complementizer, topicalization out of embedded finite clauses is degraded in Standard German. However, extraction out of non-finite embedded clauses is fine, whether or not the embedded clause has extraposed. There is no intonation break (or comma) between fronted element and matrix finite verb.

(4) a. [Dieses Buch]$_i$ habe ich [PRO den Kindern t$_i$ zu geben] versucht
    this book (ACC) have I the children (DAT) to give tried
    This book I have tried to give the children

   b. [Dieses Buch]$_i$ habe ich versucht, [PRO den Kindern t$_i$ zu geben]

### Scrambling

Scrambling in German is the movement of arguments (nominal or clausal) within the MF (rather than into the VF). The following example is from (Haider, 1991).

(5) ... daß [eine hiesige Firma]    [meinem Onkel] [die Möbel]    [vor drei Tagen]
    ... that a local company (NOM) my uncle (DAT) the furniture (ACC) three days ago
    [ohne Voranmeldung]    zugestellt hat
    without advance warning delivered has

    ...that a local company delivered the furniture to my uncle three days ago without advance warning

As Haider points out, "any permutation of these five elements (5! = 120) is grammatically well-ordered". We conclude that scrambling of more than one element is possible.

If there are embedded clauses, scrambling can move elements out of the embedded clauses ("long-distance" scrambling). However, in German, scrambling can never proceed out of tensed clauses. It has been suggested that embedded infinitival clauses form a "clause union" (Evers, 1975); if this is the case, then there is no long-distance scrambling in German because no clause boundary is crossed. However, the clause-union analysis has not gone uncontested (Kroch and Santorini, 1991). For the sake of the development in this paper, it is irrelevant whether clause union takes place and whether a clause boundary is actually crossed – the important fact is that an argument or adjunct can scramble into the domain of an (originally) different predicate. We will continue to refer to this effect as long-distance scrambling.

(6) a. ...daß niemand    [PRO den Kühlschrank zu reparieren] versprochen hat
       ...that no-one (NOM) the refrigerator (ACC) to repair    promised    has
       ...that no-one has promised to repair the refrigerator

    b. ...daß [den Kühlschrank]$_i$ niemand [$t_i$ zu reparieren] versprochen hat

There is no bound on the number of clause boundaries over which an element can scramble:

(7) a. ... daß [den Kühlschrank]$_i$    niemand    [[$t_i$ zu reparieren] zu versuchen] versprochen hat
       ... that the refrigerator (ACC) no-one (NOM)    to repair    to try    promised    has

       ...that no-one has promised to try to repair the refrigerator

    b. ... daß [den Kühlschrank]$_i$    niemand    [[$t_i$ zu reparieren] zu versuchen] zu versprechen bereit
       ... that the refrigerator (ACC) no-one (NOM)    to repair    to try    to promise    ready
       ist
       is

       ...that no-one is ready to promise to try to repair the refrigerator

Furthermore, an element scrambled (long-distance or not) from one clause does not preclude an element from another clause from being scrambled, and scrambling does not obey a "path containment" condition (Pesetsky, 1982), which would require that dependencies between moved element and trace are nested, but not crossed:

(8) ... daß [dem Kunden]$_i$    [den Kühlschrank]$_j$    bisher noch    niemand    $t_i$ [[$t_j$ zu reparieren]
    ... that the client (DAT) the refrigerator (ACC) so far as yet no-one (NOM)    to repair
    zu versuchen] versprochen hat
    to try    promised    has

    ...that so far, no-one yet has promised to repair the refrigerator

We conclude that scrambling in German is "doubly unbounded" in the sense that neither is there a bound in the competence syntax on the distance over which each element can scramble, nor is there a bound in the competence syntax on the number of unbounded dependencies that can occur in one sentence. This generalization should not be taken to mean that all sentences in which "doubly unbounded" scrambling has occurred will be judged equally acceptable. Clearly, scrambling is constrained by pragmatic and processing factors, and perhaps also by semantic factors.[3] In this paper, we will propose a competence model that allows doubly unbounded scrambling, and an associated processing model that predicts the degree of acceptability of scrambled sentences.

---

[3]Analyses of the pragmatic and semantic issues involved in scrambling (see, e.g., (Lenerz, 1977; Höhle, 1982; Moltmann, 1990)) are still somewhat sketchy; however, while they provide constraints on word order, they do not provide evidence that the generalization of "double unboundedness" must be abandoned: the contextual and semantic restrictions on word order do not translate into general rules that would categorically rule out certain formally definable orders (such as, say, word orders derived by multiple long-distance scrambling), irrespective of the particular choice of lexemes and context.

While processing load appears to increase with an increasing number of scrambled elements, the increase in processing load is gradual, providing us with a range of acceptability judgments. This picture is further complicated by the fact that long-distance scrambling, which degrades acceptability, can interact with extraposition, which improves acceptability:

(9) a. (ok) ... daß niemand       den Kühlschrank      zu reparieren zu versuchen versprochen hat
        ... that no-one (NOM) the refrigerator (ACC) to repair      to try          promised    has

   ... that no-one has promised to repair the refrigerator

b. ok ... daß niemand versprochen hat, zu versuchen, den Kühlschrank zu reparieren

c. ok ... daß niemand versprochen hat, den Kühlschrank zu reparieren zu versuchen

d. ? ... daß niemand den Kühlschrank zu versuchen zu reparieren versprochen hat

e. ?? ... daß den Kühlschrank niemand zu reparieren zu versuchen versprochen hat

(9a) is prescriptively grammatical, and native-speaker judgment can therefore not serve as an indication of processing difficulty (this is indicated by the parentheses around the judgment). However, it is well known that center-embedding presents processing difficulties. (9b) is the much preferred and perfectly acceptable fully extraposed word order. The other example sentences represent variations in decreasing order of acceptability. (9d) and (9e) include long-distance scrambling, which degrades them further.

In the case of three levels of embedding, with two of the clauses having one overt argument each, and the third clause having none (as in (9) above), we end up with 30 possible word orders. Ten of these are ruled out straightforwardly by a linguistic account (with only minimal assumptions about phrase structure, these word orders would necessarily result in unbound traces). The remaining 20 sentences display acceptabilities ranging from "perfectly acceptable" to "flat out". This "grey zone" of acceptability has not, to our knowledge, been investigated in either the linguistic or the psycholinguistic literature. This is the first phenomenon that we would like our processing model to account for.

## Topicalization and Scrambling: Two Distinct Types of Movement

We have seen that German has two distinct types of (leftward) movement: topicalization, which is movement of a single constituent (argument/adjunct) into the VF, and scrambling, which is movement of any number of constituents within the MF. There are formal differences between these two types of movement: it is known that topicalization can be handled by a simple TAG, while scrambling is beyond the formal power of TAGs (we will discuss this later on in more detail). The question arises whether this formal difference is accompanied by any linguistic differences between the two types of movement. There is evidence that the two types of movement do in fact have different linguistic properties, in particular with respect to the binding possibilities from the surface position (see Webelhuth (1989) for anaphor binding, and Frank, Lee, and Rambow (1992) for bound-variable and Principle C binding facts). Interestingly, the two types of word-order variation also appear to have different effects on the processor. Consider the following minimal pair:

(10) a. Sentence with long-distance scrambling:

   ? Der Meister   hat  den Kühlschrank       niemandem      zu reparieren   versprochen
     the master   has  the refrigerator (ACC) no-one (DAT)   to repair       promised
     The master has promised no-one to repair the refrigerator

b. Sentence with long-distance topicalization:

   Den Kühlschrank         hat   der Meister       niemandem       zu reparieren   versprochen
   the refrigerator (ACC)  has   the master (NOM)  no-one (DAT)    to repair       promised
   The master has promised no-one to repair the refrigerator

In both sentences, the argument *den Kühlschrank* 'the refrigerator' of the embedded clause has moved out of the clause and to the left. In the scrambled sentence, the surface position is just after the finite verb; in the topicalized sentence, the landing site is just before the finite verb. Nonetheless, (10b) is significantly more acceptable than (10a). We propose that this difference in acceptability is due to processing constraints, and it is the second phenomenon that we want our processing model to be able to predict.

# TAGS AND BEPDAS

The processing model that we present in this paper is based on (Joshi, 1990). Joshi (1990) proposes to model human sentence processing with a formal automaton called the Embedded Pushdown Automaton (EPDA). The EPDA is equivalent to Tree Adjoining Grammar (TAG), in the sense that for every TAG there is an EPDA that accepts exactly the set of strings that the TAG generates, and for every EPDA, there is a TAG that generates exactly the set of strings that the EPDA accepts. In this paper, we will use the bottom-up variant of the EPDA, called BEPDA,[4] which is also formally equivalent to TAG. This formal equivalence is crucial to our point: TAG has been used for the representation of competence syntax, and we propose that through the formal equivalence we can relate in a motivated manner formal models of competence directly to formal models of performance. We will therefore carefully introduce the formal notions that we will need for the remainder of our exposition. We will do so using abstract formal language examples, since linguistic examples might obscure the underlying formal mechanisms. We return to linguistic facts in the next section.

We will start out by describing TAG, which underlies our model. We then describe the automaton, and proceed to show how an automaton can be derived from a grammar.

## Context-Free Grammar and Tree Adjoining Grammar

We will first briefly review context-free grammars (CFG). While they have been all but abandoned as a basis for linguistic description in the linguistic and computational literature, they are quite familiar as a formalism and therefore useful as a starting point for the exposition of TAGs. Recall that in a CFG, we have string rewriting rules that specify how a single symbol, called a nonterminal, can be rewritten as a sequence of other symbols. We start out with a special nonterminal symbol, say $S$ for "sentence", and successively apply string rewriting rules until we have no more nonterminal symbols left in the string, but only terminal symbols (such as *John likes Mary*). (In the process, we create a *derivation* tree which records how we rewrite each nonterminal symbol: the symbols we replaced it with appear as its daughters.)

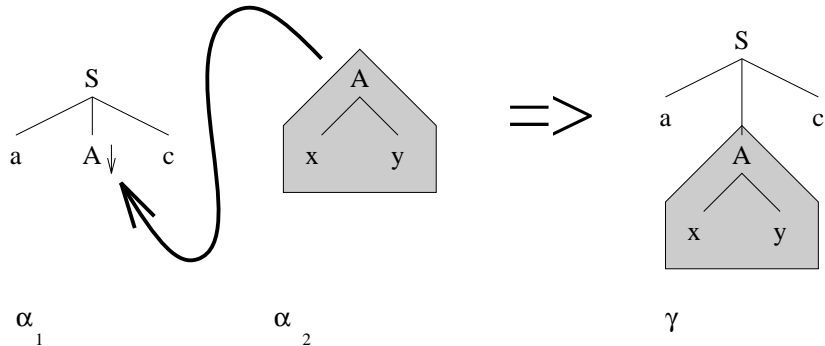

Figure 1: The Substitution Operation

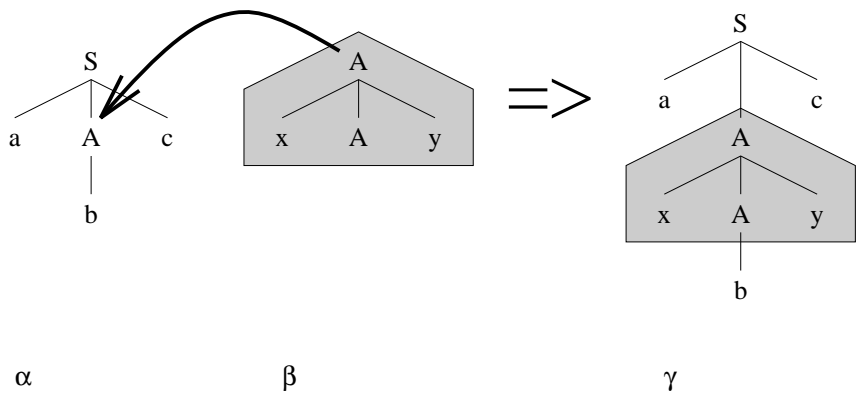

Figure 2: The Adjunction Operation

---

[4]We would like to thank Yves Schabes for suggesting the use of the BEPDA (rather than the EPDA) in modeling human sentence processing.

Just as CFG is a string rewriting system, TAG is a tree rewriting system: we start with elementary trees, and then can replace nonterminal nodes in the tree with entire trees. A TAG consists of a set of such elementary trees. Two tree combining operations are used to derive larger trees: substitution and adjunction. Substitution is shown in Figure 1: tree $\alpha_2$ can be substituted into tree $\alpha_1$ if the root node of $\alpha_2$ has the same label as a non-terminal node on the frontier of $\alpha_1$ which has been specially marked for substitution (a "substitution node"; substitution nodes are marked with down-arrows ($\downarrow$)). Adjunction is shown in Figure 2. Tree $\alpha$ (called an "initial tree") contains a non-terminal node labeled $A$; the root node of tree $\beta$ (an "auxiliary tree") is also labeled $A$, as is exactly one non-terminal node on its frontier (the "foot node"). All other frontier nodes are terminal nodes or substitution nodes. We take tree $\alpha$ and remove the subtree rooted at its node $A$, insert in its stead tree $\beta$, and then add at the footnode of $\beta$ the subtree of $\alpha$ that we removed earlier. The result is tree $\gamma$. As we can see, substitution rewrites a node on the frontier, while adjunction can rewrite an interior node, thus having the effect of inserting one tree into the center of another. For a more extensive introduction to TAGs, see Joshi (1987) and (Joshi, Vijay-Shanker, and Weir, 1991).

TAG is an appealing formalism for the representation of linguistic competence because it allows local dependencies (in particular the subcategorization frame and *wh*-dependencies) to be stated on the elementary structures of the grammar and to be factored apart from the expression of recursion and unbounded dependencies. This in turn allows us to develop a lexicon-oriented theory of syntax: because the entire subcategorization frame of a lexical item can be represented in a single tree, we can "lexicalize" the grammar in the sense that every tree is associated with exactly one lexical item (be it a word or a multi-word phraseme), and *v.v.*. It is this lexicalized version that has been used in the development of TAG grammars for English, French, German, and Japanese, and that we will be using in the remainder of this paper.

### The Push-Down Automaton and the Bottom-Up Embedded Pushdown Automaton

The Embedded Pushdown Automaton (EPDA) was introduced by Vijay-Shanker (1987) and proven to be formally equivalent to TAG. Schabes (1990) defines a bottom-up version, called BEPDA. We will only describe the BEPDA here.

We will first briefly recall the definition of the Push-Down Automaton (PDA), which is known to recognize exactly context-free languages. A PDA[5] consists of a stack of stack symbols, an input tape with a read head, and a finite state control. The read head scans the input once from left to right. A transition relation determines the moves of the automaton, based on the current state, the input symbol being scanned, and the symbol currently on the top of the stack. Two types of moves are possible:

- The automaton can shift the input symbol onto the stack (a SHIFT move).

- The automaton can remove a number of stack symbols and replace them by a single symbol (a REDUCE move).

In either case, the automaton can transition to a new state. The automaton accepts the input if, upon reading it completely, its stack is empty. It should be noted that a PDA (in general) is non-deterministic, meaning that for a given state, input symbol, and top of stack, several different moves are possible. Given a context-free grammar, it is easy to construct a corresponding PDA (though it will not be the only one that corresponds to that CFG): we simply need to SHIFT any non-terminal from the input tape onto the stack and to interpret the context-free rules as REDUCE moves (if the right-hand side of a rule is on the top of the stack, it REDUCEs to the left-hand side).

Now let us turn to the BEPDA. Like the PDA, the BEPDA consists of a push-down store, an input tape with a one-way read-only scanner, and a finite state automaton that controls the actions of the automaton. The push-down store has a more complex structure than that of a PDA: it is a stack of stacks of stack symbols, rather than a simple stack of stack symbols. A transition relation is defined for the automaton between triples consisting of the current state, the input symbol and the stack symbol on the top of the top stack on the one hand and pairs consisting of a new state and new material to be put on the pushdown store on the other hand. The actions are as follows:

- The SHIFT move first creates a new (empty) stack on the top of the stack of stacks and then pushes a single symbol onto it.

---

[5]We give a particular definition of PDA; other equivalent ones are possible.

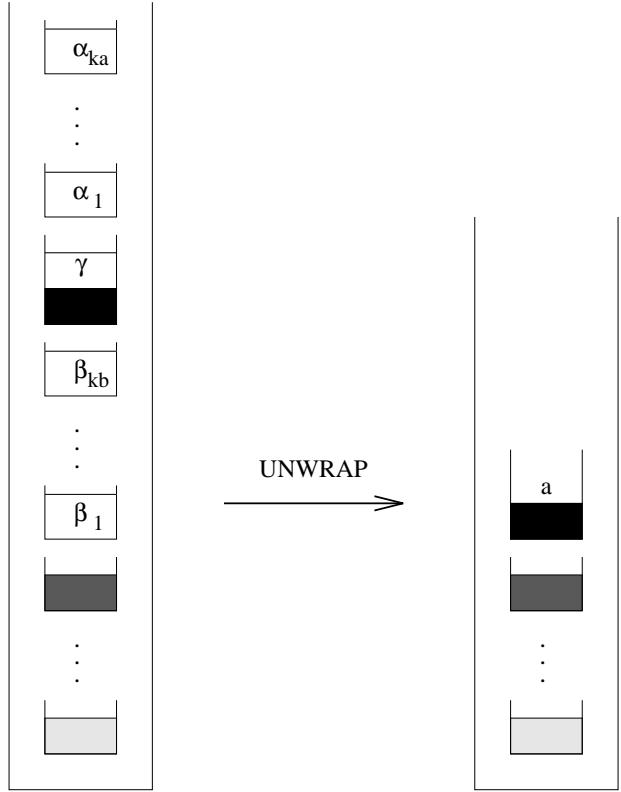

Figure 3: UNWRAP Move of the BEPDA

- The UNWRAP operation is an extension of the PDA REDUCE move. This operation is illustrated in Figure 3. In an UNWRAP move, finite (and possibly empty) sequences of stacks are removed immediately below (stacks $\beta_1, \ldots, \beta_{k_b}$ in the figure) and above (stacks $\alpha_1, \ldots, \alpha_{k_a}$) a designated stack (which becomes the new top stack). (We say that these stacks are "UNWRAPped around" the new top stack.) Then, a sequence of (possibly empty) stack symbols on the new top stack is popped ($\gamma$ in the figure) and replaced by a single new stack symbol (a in the figure).

There are several "degenerate" UNWRAP moves that are used often and therefore have special names. They are degenerate in the sense that some of the stack symbols involved in the definition of the UNWRAP move are specified to be the empty string. For example, if the stacks below and above the new top stack are empty (i.e., the new top stack is the old top stack), and if furthermore the new stack symbol is also empty, we have simply removed stack symbols from the top of the top stack. We will call this move a POP. If, however, if in addition to the stacks below and above the new top stack $\gamma$ is also empty, i.e., we do not remove any stack symbols at all, but just push a new stack symbol on to the top stack, then we will call the move a PUSH move.

It is the UNWRAP move that extends the power of the BEPDA beyond that of the PDA and its REDUCE move: the stacks removed above and below the new top stack correspond to adjoined material. We will illustrate this by giving an automaton for the sample tree adjoining grammar of Figure 2.

The automaton has a single state, $q$. We then have the following transition rules:

1. Any scanned input symbol ($a$, $b$, $c$, $x$, or $y$) is SHIFTed onto the stack of stacks.

2. A top top-stack symbol of $b$ is replaced by $A_\alpha$ (a degenerate form of UNWRAP).

3. If the top top-stack symbol is $A_\alpha$, $A_\beta$ may be PUSHed on top of it.

4. If the top stack consists of the element $y$ (and nothing else), the third stack of the element $x$ (and nothing else), and if the top of the second stack is $A_\beta$, then $x$, $A_\beta$, and $y$ can be UNWRAPped around the second stack, and another copy of $A_\beta$ pushed onto it.

5. $A_\beta$ on the top of the top-stack can be POPped.

6. If the top stack consists of the element $c$ (and nothing else), the third stack of the element $a$ (and nothing else), and if the top of the second stack is $A_\alpha$, then $x$, $A_\alpha$, and $y$ can be UNWRAPped around the second stack, and $S$ pushed onto it.

7. $S$ on the top of the top-stack can be POPped.

For every TAG, there are many different equivalent BEPDAs. We have constructed this BEPDA in a specific manner, which is particularly straightforward in that it establishes a close relation between the grammar and the automaton. We will call this construction the "simple method". First, we have distinguished nodes in different trees that bear the same label by subscripts. The stack symbols are just the terminal symbols and the nonterminal nodes (identified by their label and the tree index) from the trees of the grammars, so we will speak (somewhat sloppily) of a tree node being in the pushdown store. We have then constructed the rules as follows. Apart from the SHIFT moves (rule (1)), we have "exploded" the trees in the TAG into a set of context-free rules, each describing a node and its daughters. We have then associated one UNWRAP move with each of these context-free rules (rule (4) for tree $\alpha$ and rule (6) for tree $\beta$). Furthermore, at each node at which an adjunction is possible, we have added a rule that PUSHes the footnode of each adjoinable tree onto that node. In our example, the only possible adjunction is of tree $\beta$ into tree $\alpha$ at the node labeled $A$ (rule (3)). Finally, any root node of a tree can be POPped off the automaton (rule (5) for tree $\alpha$ and rule (7) for tree $\beta$). We will now present a run of this automaton, as it accepts the input string $axbyc$. The pushdown store is shown "sideways", with the top to the right. The symbol '[' denotes the bottom of a stack.

(11)

| Step | In State | Store | | | | Input Consumed | Rule Used |
|------|----------|-------|---|---|---|----------------|-----------|
| 0 | $q$ | | | | | | |
| 1 | $q$ | [$a$ | | | | $a$ | (1) |
| 2 | $q$ | [$a$ | [$x$ | | | $x$ | (1) |
| 3 | $q$ | [$a$ | [$x$ | [$b$ | | $b$ | (1) |
| 4 | $q$ | [$a$ | [$x$ | [$A_\alpha$ | | — | (2) |
| 5 | $q$ | [$a$ | [$x$ | [$A_\alpha\ A_\beta$ | | — | (3) |
| 6 | $q$ | [$a$ | [$x$ | [$A_\alpha\ A_\beta$ | [$y$ | $y$ | (1) |
| 7 | $q$ | [$a$ | [$A_\alpha\ A_\beta$ | | | — | (4) |
| 8 | $q$ | [$a$ | [$A_\alpha$ | | | — | (5) |
| 9 | $q$ | [$a$ | [$A_\alpha$ | [$c$ | | $c$ | (1) |
| 10 | $q$ | [$S$ | | | | — | (6) |
| 11 | $q$ | | | | | — | (7) |

We observe two crucial points about how the automaton recognizes the adjoined tree. First, the automaton "decides" to start simulating an adjunction in step 5 by PUSHing the footnode of the auxiliary tree onto the top stack. Now the automaton proceeds in steps 6 and 7 by recognizing the auxiliary tree $\beta$ completely independently from the initial tree $\alpha$ into which it is adjoined. The only sign that a tree $\alpha$ has been partially recognized is the $A_\alpha$ node below the $A_\beta$ node in the stack (at the top of the stack of stacks after steps 5 and 7, the second from the top after step 6). But since $A_\alpha$ is not at the top of any stack, it does not affect the processing of the automaton during the recognition of tree $\beta$. Here it becomes apparent why we need a stack of *stacks* (because we need to store information about partially recognized trees) and why we need an UNWRAP move (the automaton needs to manipulate the top of stacks without affecting the information stored in them below). Second, once tree $\beta$ has been fully recognized, its root node is at the top of the top stack. In step 8 (after recognizing the $y$) the node is POPped from the pushdown store. Thereafter, there is no trace at all of tree $\beta$ ever having been recognized; the automaton continues just as if it were simply recognizing the string $abc$, whose derivation requires no adjunction.

## A FORMAL MODEL OF SYNTACTIC PROCESSING

In this section, we use the machinery introduced in the previous section to model human syntactic processing. We first give examples of how processing models are derived from competence grammars, and then we show how the automaton can be equipped with a metric to predict processing load.

### Linguistic TAGs and associated EPDAs

As we have previously pointed out, the formal equivalence between TAG and BEPDA means that for every tree adjoining language $L$, there is at least one BEPDA $M$ that recognizes exactly $L$, and typically there is an infinite number of such automata. However, the existence of *some* automaton that makes psycholinguistically relevant predictions is not of interest unless we know how to choose the right automaton from among those that are formally equivalent (but not all of which make the right predictions). We propose that the simple method introduced in the previous section will derive the right automaton. From a linguistically motivated grammar it will derive an automaton that models syntactic processing in a plausible way. This approach shifts the problem of providing a principled account of how to construct models of the syntactic processor (i.e., automata) to the problem of how to construct (competence) grammars – which is of course the object of syntax. This point is important because it affects the question of the universality of the parser. The processor behaves differently for different languages. If we want to explain this variation in a principle-and-parameter type methodology, we have two options: either we can assume that the processor is parametrized on its own (though the parameter setting may be linked to the setting of the linguistic parameters), or we can deduce the cross-linguistic variation among processors from cross-linguistic variation among competence grammars and the way in which the (universal) processor interacts with the competence grammar. The latter view is adopted by Inoue and Fodor (1993), who term it "as-if-parametrized". Our approach falls into this paradigm as well.

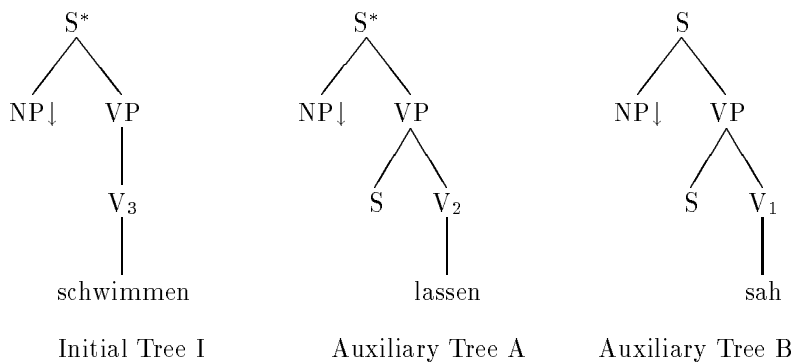

Figure 4: The German grammar

We now turn to linguistic grammars and show how the simple method derives automata. Let us consider the following grammar for a fragment of German given in Figure 4. In this grammar, matrix clauses are adjoined into their subordinate clauses at the root S node. This analysis is motivated by facts about *wh*-extraction out of subordinate clauses as discussed by Kroch (1987; 1989): if we assume that the *wh*-word is included in the same tree as its governing verb (at an S' node) and adjoin the matrix clause at the S node, then we get subjacency effects "for free". Further cross-linguistic evidence for this analysis of clausal embedding comes from Dutch. Kroch and Santorini (1991) give extensive syntactic evidence for the grammar given in Figure 6. Again, the correct derivation of the cross-serial dependencies relies on the adjunction of the matrix clause into its subordinate. We therefore adopt this approach. Note that it is motivated by purely linguistic considerations – no processing issues intervened in the formulation of this grammar. The grammar in Figure 4 can generate center-embedded sentences such as the following:

(12)  ... daß  Peter  Maria  die Kinder       schwimmen  lassen   sah
       ... that  Peter  Maria  the children (ACC)  swim (inf)  let (inf)  saw
            $N_1$    $N_2$    $N_3$              $V_3$          $V_2$       $V_1$

       ...that Peter sah Maria let the children swim

German sentence (12) is derived by first substituting the nominal arguments into the NP substitution nodes of trees I, A, and B. Then, auxiliary tree A is adjoined into initial tree I at the root node of I, and auxiliary tree B is adjoined into the root node of the derived tree (which is in fact the root node of auxiliary tree A). The resulting structure, the derived tree, is shown in Figure 5.

We now construct an automaton using the simple method. We start out by "exploding" the trees into context-free rules. For Tree I, we obtain the following context-free rules:

(13) Derived context-free rules for Tree I in Figure 4:

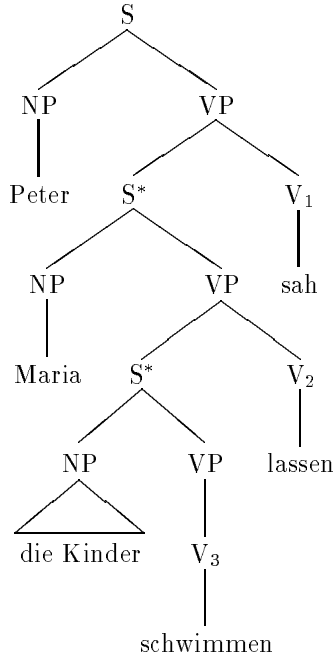

Figure 5: German derived tree

$S_I \longrightarrow NP_I\ VP_I$
$VP_I \longrightarrow V_I$
$V_I \longrightarrow schwimmen$

As we have stressed during the discussion of the TAG formalism, the elementary structures of a TAG are trees, and therefore we can associate every elementary structure with a single lexical item ("lexicalization"). Since the automaton mimics the derivation in a TAG, we can interpret the moves of the BEPDA as establishing connections between lexical items. This view of the processor – as manipulating lexical items, and not phrase structure nodes – is somewhat different from the view on processing that arises when one starts from context-free grammars. In order to emphasize the lexical orientation of the BEPDA model, we will follow Joshi (1990) in giving the stack symbols quasi-categorial labels, rather than the standard labels. Suppose we have a clausal tree (anchored on a verb) that has substitution nodes $\alpha_1, \ldots, \alpha_n$. Recall that substitution nodes are nodes at which substitution must occur in order to make the tree a complete initial or auxiliary tree. Linguistically, they correspond to nominal arguments. We will associate with the verb and each of the nodes in its projection (V, VP and S in this case) a label of the form $V\{\alpha_1, \ldots, \alpha_k\}$, where the $\alpha_i$ are labels of substitution nodes that have not yet been filled. Thus, linguistically, the list represents unfulfilled nominal subcategorization requirements. It is important to note that this notation is entirely equivalent to the notation used in writing the trees themselves: in our sample trees, the VP corresponds to V{NP}, and the top S node to V{}. The rules of the BEPDA that we obtain are as follows.

(14) BEPDA rules derived from Tree I in Figure 4:

1. SHIFT *die Kinder* and *schwimmen* onto the pushdown store.
2. UNWRAP [*schwimmen* and replace by [$V_I${NP}
3. UNWRAP [NP around [$V_I${NP} and replace by [$V_I${}
4. POP [$V_I${}

The first rule in (14) implements the principle of the simple method to push any scanned input symbols onto the pushdown store. Rule 2 represents the projection of *schwimmen* to the VP and corresponds to the second and third context-free rules from (13). In the third rule, the subject and the VP are combined to form the sentence, corresponding to the first context-free rule of (13). Finally, the fourth rule implements the principle that root nodes of fully recognized trees are removed from the automaton. Note that we assume that all NP symbols (both stack symbols and those in the subcategorization set) are marked with case information. Furthermore, we assume that we have a syntax of NPs that will allow us to recognize full NPs; we omit the details.

The BEPDA is defined as a non-deterministic automaton. This is not appealing as a model of human sentence processing. We observe that the notion of incremental processing means that the syntactic processor performs as much computation as it can on a given input token, rather than wait for the complete sentence before processing initial parts of it. We will therefore assume the following ordering principle on the application of BEPDA rules:

(15) **Ordering Principle for BEPDA rules:**

1. Perform all possible UNWRAP and POP moves first.
2. SHIFT a new input item only when no further UNWRAP moves are possible.

This of course does not address the problem that arises when two UNWRAP moves are possible, which represents cases of true syntactic ambiguity such as PP-attachment. Such syntactic ambiguity does not arise in the cases we are interested in in this paper, and the automaton does not immediately make predictions about preferences. We return briefly to the issue of syntactic ambiguity in the conclusion.

As an example, suppose we want to use the BEPDA to recognize the sentence fragment *die Kinder schwimmen*. The run of the automaton would be as follows:

(16)

| Step | Rule Applied | Pushdown Store Configuration |
|---|---|---|
| 1 | SHIFT *die Kinder* (Rule 1) | [*die Kinder* |
| 2 | Use NP BEPDA rules to derive | [NP |
| 3 | SHIFT *schwimmen* (Rule 1) | [NP  [*schwimmen* |
| 4 | UNWRAP *schwimmen* (Rule 2) | [NP  [$V_I${NP} |
| 5 | UNWRAP NP (Rule 3) | [$V_I$ |
| 6 | POP (Rule 4) | |

The empty pushdown store at the end indicates a successful recognition of a clausal unit. Since the rules derived from the grammar in Figure 4 are structurally similar, we can group them together into the following table:

(17) **Automaton for German:**

| State | Read Head Scans | Top of Stack is | Action |
|---|---|---|---|
| 1 | N | *anything* | SHIFT N |
| 2 | V | *anything* | SHIFT V |
| 3 | *anything* | [N [V | UNWRAP N around V |
| 4 | *anything* | [V [V' | UNWRAP V' around V |

In the third rule, V must have an unfulfilled subcategorization requirement for a noun (with matching case features), and in the fourth rule, V' must have an unfulfilled subcategorization requirement for a clause. Now let us turn to recognition of German center-embedded sentence (12). In (18), the column "Input" shows the symbol being scanned by the read head.

(18)

| Step | In State | Store | | | | Input Read | Input Scanned |
|------|----------|-------|---|---|---|------------|---------------|
| 0 | 0 | | | | | | |
| 1 | 1 | [$N_1$ | | | | $N_1$ | $N_1$ |
| 2 | 1 | [$N_1$ | [$N_2$ | | | $N_2$ | $N_2$ |
| 3 | 1 | [$N_1$ | [$N_2$ | [$N_3$ | | $N_3$ | $N_3$ |
| 4a | 2 | [$N_1$ | [$N_2$ | [$N_3$ | [$V_3\{N_3\}$ | $V_3$ | $V_3$ |
| 4b | 2 | [$N_1$ | [$N_2$ | [$V_3\{\}$ | | — | $V_2$ |
| 5a | 2 | [$N_1$ | [$N_2$ | [$V_3\{\}$ | [$V_2\{N_2\}$ | $V_2$ | $V_2$ |
| 5b | 2 | [$N_1$ | [$N_2$ | [$V_3\{\}$ $V_2\{N_2\}$ | | — | $V_1$ |
| 5c | 2 | [$N_1$ | [$V_3\{\}$ $V_2\{\}$ | | | — | $V_1$ |
| 6a | 2 | [$N_1$ | [$V_3\{\}$ $V_2\{\}$ | [$V_1\{N_1\}$ | | $V_1$ | $V_1$ |
| 6b | 2 | [$N_1$ | [$V_3\{\}$ $V_2\{\}$ $V_1\{N_1\}$ | | | — | — |
| 6c | 2 | [$V_3\{\}$ $V_2\{\}$ $V_1\{\}$ | | | | — | — |
| 6d | 3 | [$V_3\{\}$ $V_2\{\}$ | | | | — | — |
| 6e | 3 | [$V_3\{\}$ | | | | — | — |
| 6f | 3 | | | | | | |

In steps one through three, the NPs are recognized and stored in separate stacks in the pushdown store. In step 4a, a verb is read in. Since the case features match, the UNWRAP rule derived from tree I applies and the nominal subcategorization requirement of $V_3$ is fulfilled. Note that we have now recognized tree I to the root node. However, it cannot be removed from the automaton, since a further adjunction at the root node (in terms of the automaton, a further UNWRAP) may occur, and in fact does occur in our example. We therefore read in the next input $V_2$ (step 5a), which can be UNWRAPped around the top stack (step 5b). After step 5c, the top stack contains $V_2\{\}$ on top of $V_3\{\}$, representing the fact that we have recognized $V_2$'s tree A (with its nominal argument substituted) adjoined into $V_3$'s tree I (with its nominal argument substituted). Finally, in steps 6a and 6b, we read in $V_1$, and UNWRAP it and $N_1$ around the top stack. We can now pop $V_1\{\}$, followed by $V_2\{\}$, and finally $V_3\{\}$.

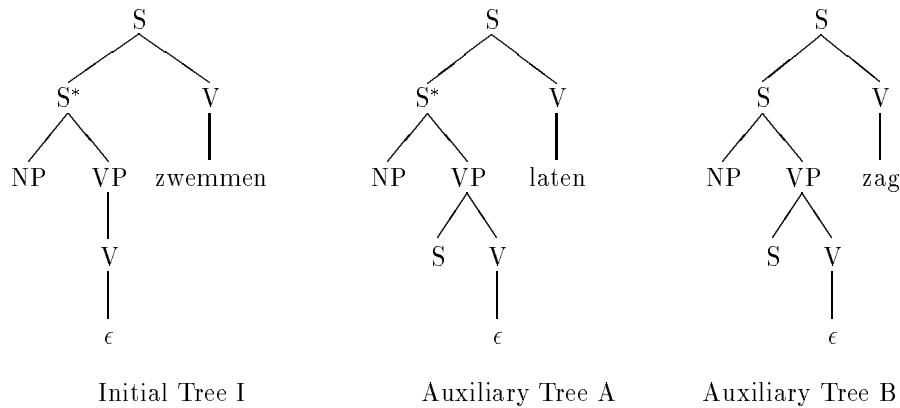

Figure 6: The Dutch grammar

We will now very briefly discuss Dutch. We use the grammar for Dutch given in Kroch and Santorini (1991), repeated in Figure 6. Kroch and Santorini give extensive syntactic evidence for their grammar, but do not consider processing issues at all. The Dutch cross-serial dependencies are derived by adjoining Tree B into Tree A at the node marked $S^*$, and then adjoining the combination into Tree I, again at the node marked $S^*$. Note that the cross-serial dependencies are a result only of head movement (verb raising) that has occurred locally in each clause, not of multi-clausal ordering rules or verb-complex formation.

We derive an automaton in an analogous manner to the German case. There is, however, one complication: the empty category that results from head movement of the verb. Since the bottom-up recognition of a tree can only proceed once the empty head has been posited, the automaton must have a rule for hypothesizing empty heads. Clearly, we do not want it to do so non-deterministically, and we must extend our algorithm for deriving automata from competence grammars. We do so by defining the following two conditions on processing empty heads, or, more precisely, subtrees whose yield is the null string and which contain a head trace ("headed null-subtrees"). Both conditions must be met.

- **Bottom-up condition:** The automaton assumes the recognition of a headed null-subtree only if it has recognized its sister subtree and it can perform an UNWRAP operation involving the headed null-subtree.

- **Top-down condition:** A headed null-subtree can only be posited if it is licensed top-down by a previously processed licensor (in which case the licensing relationship must be indicated on the top of the push-down store by features), or if the input symbol the read head is currently scanning licenses an empty head of the appropriate type.

We will denote the projection from a (verbal) head trace by $V^{h-}$, while we will use $V^{h+}$ to indicate an overt full lexical head. With this approach, our Dutch automaton looks as follows (we omit the details of the construction):

(19) **Automaton for Dutch:**

| State | Read Head | Top of Stack | Action |
|---|---|---|---|
| 1 | N | *anything* | SHIFT N |
| 2 | $V^{h+}$ | *anything* | Assume $V^{h-}$ (an empty V head projection) and UNWRAP |
| 3 | $V^{h+}$ | $V^{h-}$ | SHIFT $V^{h+}$ and UNWRAP |

Recall that all possible UNWRAP moves are performed before any possible SHIFT moves.

Consider the following sentence:

(20)  ... omdat    Piet    Marie    de kinderen    zag    laten    zwemmen
      ... because  Piet    Marie    the children   saw    let      swim
                   $N_1$   $N_2$    $N_3$          $V_1^{h+}$  $V_2^{h+}$  $V_3^{h+}$
      ...because Piet saw Marie let the children swim

Given the input $N_1$ $N_2$ $N_3$ $V_1^{h+}$ $V_2^{h+}$ $V_3^{h+}$, the automaton executes the following steps:

(21)

| Step | In State | Store | | | Input Read | Input Scanned |
|---|---|---|---|---|---|---|
| 0 | 0 | | | | | |
| 1 | 1 | $[N_1$ | | | $N_1$ | $N_1$ |
| 2 | 1 | $[N_1$ | $[N_2$ | | $N_2$ | $N_2$ |
| 3 | 1 | $[N_1$ | $[N_2$ | $[N_3$ | $N_3$ | $N_3$ |
| 4a | 2 | $[N_1$ | $[N_2$ | $[V_3^{h-}\{\}$ | — | $V_1^{h+}$ |
| 4b | 2 | $[N_1$ | $[N_2$ | $[V_3^{h-}\{\}\ V_2^{h-}\{N_2\}$ | — | $V_1^{h+}$ |
| 4c | 2 | $[N_1$ | $[V_3^{h-}\{\}\ V_2^{h-}\{\}$ | | — | $V_1^{h+}$ |
| 4d | 2 | $[N_1$ | $[V_3^{h-}\{\}\ V_2^{h-}\{\}\ V_1^{h-}\{N_1\}$ | | — | $V_1^{h+}$ |
| 4e | 2 | $[V_3^{h-}\{\}\ V_2^{h-}\{\}\ V_1^{h-}\{\}$ | | | — | $V_1^{h+}$ |
| 5a | 2 | $[V_3^{h-}\{\}\ V_2^{h-}\{\}\ V_1^{h-}\{\}$ | $[V_1^{h+}$ | | $V_1^{h+}$ | $V_1^{h+}$ |
| 5b | 2 | $[V_3^{h-}\{\}\ V_2^{h-}\{\}\ V_1^{h+}\{\}$ | | | — | $V_2^{h+}$ |
| 5c | 2 | $[V_3^{h-}\{\}\ V_2^{h-}\{\}$ | | | — | $V_2^{h+}$ |
| 6a | 2 | $[V_3^{h-}\{\}\ V_2^{h-}\{\}$ | $[V_2^{h+}$ | | $V_2^{h+}$ | $V_2^{h+}$ |
| 6b | 2 | $[V_3^{h-}\{\}\ V_2^{h+}\{\}$ | | | — | $V_3^{h+}$ |
| 6c | 2 | $[V_3^{h-}\{\}$ | | | — | $V_3^{h+}$ |
| 7a | 2 | $[V_3^{h-}\{\}$ | $[V_3^{h+}$ | | $V_3^{h+}$ | $V_3^{h+}$ |
| 7b | 2 | $[V_3^{h+}\{\}$ | | | — | — |
| 7c | 2 | | | | — | — |

First, the three nouns are read into the push-down store on separate stacks. Before step 1, the bottom-up condition for positing a headed null-subtree is not met. Before steps 2 and 3, the bottom-up condition is met, but not the top-down condition: no licensing verbal head has been read, nor is the input head scanning a potential licensor. The latter condition is met after step 3, so that in steps 4a through 4e headed null-subtrees are posited and the appropriate UNWRAP moves are performed. At the end of step 4, the empty heads with saturated subcategorization requirements are stacked on one stack, representing the fact that tree A is adjoined into tree I, and tree B into tree A, with the adjunctions taking place below the lexical verb but above its trace. Then, in steps 5, 6, and 7, the lexical heads are read in, UNWRAPped with the top of the stack, and the completed structures are popped off the push-down store.

## Measuring Processing Load

Bach, Brown, and Marslen-Wilson (1986) showed experimentally that native speakers of German take longer to process sentences with nested dependencies than native speakers of Dutch take to process equivalent sentences with cross-serial dependencies. Joshi (1990) showed how a processing model based on a (top-down) EPDA predicts these facts. In this section, we will show that the BEPDA model proposed in this paper makes the same predictions, while being derived from linguistically motivated grammars.

We will associate a metric with the run of the automaton. Joshi (1990) proposes two metrics, a simple one that simply registers the maximum number of items stored during processing of a given input sentence, and a more complex one which also takes into account how long each item spends in the push-down store. We will adopt the second approach, though we will modify it slightly. The basic idea is to record for each step how many items are stored in the push-down store (each lexical item is given a score of 1), and then sum up the scores for all steps. A step is defined as a move of the automaton, whether or not input is read.[6] For center-embedded sentences, the automaton for German gives us the following score:

---

[6]Joshi (1990) defined a step to be a move in which input is read, which explains the different values given in that paper and here. The difference between the EPDA and the BEPDA models is not relevant: if we use the original convention for scoring for the BEPDA model, we obtain exactly the same scores as Joshi (1990).

(22)

| Step | Store | | | | Input Read | New Score | Cumulative Score |
|---|---|---|---|---|---|---|---|
| 0 | | | | | | | 0 |
| 1 | [$N_1$ | | | | $N_1$ | 1 | 1 |
| 2 | [$N_1$ | [$N_2$ | | | $N_2$ | 2 | 3 |
| 3 | [$N_1$ | [$N_2$ | [$N_3$ | | $N_3$ | 3 | 6 |
| 4a | [$N_1$ | [$N_2$ | [$N_3$ | [$V_3\{N_3\}$ | $V_3$ | 4 | 10 |
| 4b | [$N_1$ | [$N_2$ | [$V_3\{\}$ | | | 4 | 14 |
| 5a | [$N_1$ | [$N_2$ | [$V_3\{\}$ | [$V_2\{N_2\}$ | $V_2$ | 5 | 19 |
| 5b | [$N_1$ | [$N_2$ | [$V_3\{\}$ $V_2\{N_2\}$ | | | 5 | 24 |
| 5c | [$N_1$ | [$V_3\{\}$ $V_2\{\}$ | | | | 5 | 29 |
| 6a | [$N_1$ | [$V_3\{\}$ $V_2\{\}$ | [$V_1\{N_1\}$ | | $V_1$ | 6 | 35 |
| 6b | [$N_1$ | [$V_3\{\}$ $V_2\{\}$ $V_1\{N_1\}$ | | | | 6 | 41 |
| 6c | [$V_3\{\}$ $V_2\{\}$ $V_1\{\}$ | | | | | 6 | 47 |
| 6d | [$V_3\{\}$ $V_2\{\}$ | | | | | 4 | 51 |
| 6e | [$V_3\{\}$ | | | | | 2 | 53 |
| 6f | | | | | | | |

"New Score" refers to the score contributed by that step, while "Cumulative Score" is, of course, the total up to and including that step. Note that a verb with its subcategorization requirements fulfilled contributes a score corresponding to the verb and its nominal arguments. Now consider the Dutch automaton:

(23)

| Step | Store | | | Input Read | New Score | Cumulative Score |
|---|---|---|---|---|---|---|
| 0 | | | | | | 0 |
| 1 | [$N_1$ | | | $N_1$ | 1 | 1 |
| 2 | [$N_1$ | [$N_2$ | | $N_2$ | 2 | 3 |
| 3a | [$N_1$ | [$N_2$ | [$N_3$ | $N_3$ | 3 | 6 |
|    | [$N_1$ | [$N_2$ | [$V_3^{h-}\{\}$ | $V_1^{h+}$ | 0 | 6 |
| 3b | [$N_1$ | [$N_2$ | [$V_3^{h-}\{\}$ $V_2^{h-}\{N_2\}$ | $V_1^{h+}$ | 3 | 9 |
| 3c | [$N_1$ | [$V_3^{h-}\{\}$ $V_2^{h-}\{\}$ | | $V_1^{h+}$ | 3 | 12 |
| 3d | [$N_1$ | [$V_3^{h-}\{\}$ $V_2^{h-}\{\}$ $V_1^{h-}\{N_1\}$ | | $V_1^{h+}$ | 3 | 15 |
| 3e | [$V_3^{h-}\{\}$ $V_2^{h-}\{\}$ $V_1^{h-}\{\}$ | | | $V_1^{h+}$ | 3 | 18 |
| 4a | [$V_3^{h-}\{\}$ $V_2^{h-}\{\}$ $V_1^{h-}\{\}$ | [$V_1^{h+}$ | | $V_1^{h+}$ | 4 | 22 |
| 4b | [$V_3^{h-}\{\}$ $V_2^{h-}\{\}$ $V_1^{h+}\{\}$ | | | $V_2^{h+}$ | 4 | 26 |
| 4c | [$V_3^{h-}\{\}$ $V_2^{h-}\{\}$ | | | $V_2^{h+}$ | 2 | 28 |
| 5a | [$V_3^{h-}\{\}$ $V_2^{h-}\{\}$ | [$V_2^{h+}$ | | $V_2^{h+}$ | 3 | 31 |
| 5b | [$V_3^{h-}\{\}$ $V_2^{h+}\{\}$ | | | $V_3^{h+}$ | 3 | 34 |
| 5c | [$V_3^{h-}\{\}$ | | | $V_3^{h+}$ | 1 | 35 |
| 6a | [$V_3^{h-}\{\}$ | [$V_3^{h+}$ | | $V_3^{h+}$ | 2 | 37 |
| 6b | [$V_3^{h+}\{\}$ | | | — | 2 | 39 |
| 6c | | | | — | 0 | 39 |

Our model is based on the assumption that the automaton manipulates representations of lexical items. This assumption dictates two conventions about scoring that we have made above. First, empty heads do not contribute to the score, since they are not associated with any lexical item. Second, the recognition of the first empty ($V_1^{h-}$) head does not count as a move, since it merely results in a relabeling of a stack symbol, and not in a change of the stack configuration. Such relabelings can also be implemented by a simple change in state, in which case no actual operation on the pushdown store is involved. Empirically, these two implementations are indistinguishable. The scores by number of clauses are as follows:

(24)

| Level of Embedding | Dutch | German |
|---|---|---|
| 1 | 5 | 5 |
| 2 | 18 | 23 |
| 3 | 39 | 53 |
| 4 | 72 | 95 |

Finally, without going into much detail, we give a run of an automaton for a German sentence with extraposition, $N_1\ V_1\ N_2\ V_2\ N_3\ V_3$. The grammar is as given above in Figure 4, except that in trees A and B, the S node follows, rather than precedes, the V node.[7]

(25)

| Step | Store | Input | New | Cum |
|---|---|---|---|---|
| 0 | | | 0 | 0 |
| 1 | $[N_1$ | $N_1$ | 1 | 1 |
| 2a | $[N_1\ [V_1\{N_1\}$ | $V_1\{N_1\}$ | 2 | 3 |
| 2b | $[V_1\{\}$ | | 2 | 5 |
| 2c | | | 0 | 5 |
| 3 | $[N_2$ | $N_2$ | 1 | 6 |
| 4a | $[N_2\ [V_2\{N_2\}$ | $V_2\{N_2\}$ | 2 | 8 |
| 4b | $[V_2\{\}$ | | 2 | 10 |
| 4c | | | 0 | 10 |
| 5 | $[N_3$ | $N_3$ | 1 | 11 |
| 6a | $[N_3\ [V_3\{N_3\}$ | $V_3\{N_3\}$ | 2 | 13 |
| 6b | $[V_3\{\}$ | | 2 | 15 |
| 6c | | | 0 | 15 |

We see that we can remove each clause as it is recognized, since we start with the matrix clause. In extraposed constructions, the scores grow linearly with the number of embedded clauses, while in center-embedded constructions, the scores grow with the square of the number of embeddings. The following table refers to German data:

(26)

| Level of Embedding | Center-Embedded | Extraposed |
|---|---|---|
| 1 | 3 | 5 |
| 2 | 23 | 10 |
| 3 | 53 | 15 |
| 4 | 95 | 20 |

The automaton model predicts strongly that extraposition is preferred over center-embedding, in particular at levels of embedding beyond two. This prediction is confirmed by native-speaker intuition, and we conjecture that psycholinguistic experiments or corpus-based studies would come to the same conclusions.

### The "Principle of Partial Interpretation"

Why does the BEPDA automaton model make different predictions for Dutch and German sentences of comparable level of embedding? The main reason is that in the Dutch sentences, clauses are removed from the push-down store as soon as the first verb is read in, while in German, clauses are only removed once the last verb of the sentence has been processed. Bach, Brown, and Marslen-Wilson (1986) interpret their experimental data as suggesting such a behavior by the processor, and suggest that this behavior arises because structures can only be removed from the processor once there is a place for them "to attach to" – an embedded clause cannot be removed while its matrix clause is still in the processor. Joshi (1990) proposes to formalize this intuition by defining a restriction on the way that automaton works, called the "Principle of Partial Interpretation" (PPI). The PPI makes the following two stipulations:

1. Structures are only discharged from the automaton when they are a properly integrated predicate-argument structure. More precisely, a clausal structure must contain all of the nominal arguments it subcategorizes for.

2. A structure is discharged only when it is either the root clause or it is the immediately embedded clause of the previously discharged structure.

---

[7]*Lassen* 'to let' does not allow extraposition. Any verb that takes a *zu*-infinitive does, and can be used in its stead.

In our discussion so far, we have not appealed to the PPI. For the types of structures under consideration, we have not needed to do so: the PPI is simply a consequence of the simple method used to derive the automaton from the competence grammar, and independently motivated ways in which the competence grammar is defined. The reason for this is that adjunction in the grammar is simulated in the automaton by recognizing the adjoined tree bottom-up and then removing any trace of it once its root node has been reached (UNWRAPping). Thus, the first material to be removed from the automaton corresponds to the last tree adjoined. Substitution, on the other hand, is handled differently: the material corresponding to the substituted tree is not removed from the automaton; in fact, it is treated as if it were part of the tree into which it was substituted. Thus, we see that the first part of the PPI follows from the fact that, in the competence grammar, we substitute nominal arguments into the trees of their governing verbs. The second part of the PPI follows from the fact that, in the competence grammar, we adjoin matrix clauses into their subordinate clauses. We intend to investigate further whether the PPI is required as an independently stated principle of processing by considering other constructions from other languages.

## HANDLING LONG-DISTANCE DEPENDENCIES

How do we handle long-distance scrambling? It has been shown formally that TAGs are not powerful enough to derive the full range of scrambled sentences (Becker, Joshi, and Rambow, 1991; Rambow, Becker, and Niv, 1992).[8] We therefore introduce a multi-component extension of TAG called VMC-TAG-DL. In multi-component TAG systems, several trees are grouped together into a set. In VMC-TAG-DL, there is no "locality" restriction on where we may adjoin trees from one set, as there are in the so-called Linear Context-Free Rewriting Systems (Weir, 1988). There is also no requirement that trees from one set be adjoined simultaneoudly, nor that the trees from one set be adjoined one immediately following the other. We can first adjoin one tree from a set, then go on and adjoin some trees from a second set, and then return to the first set to adjoin the remaining trees. The only requirement is that at the end of the derivation, either no or all trees from a given instance of a tree set must have been adjoined. Furthermore, the trees in a set are connected by *dominance links* (indicated by dotted lines in the figures). A dominance link indicates that when the derivation has terminated, the nodes linked by the dominance link must be in a relation of (not necessarily immediate) dominance. Linguistically, we use dominance links to enforce a c-command relation between related elements. Let us consider as an example the tree set for *versprechen* 'to promise', shown in Figure 7. We can think of this tree set as representing a head (the verb) and its projection (as is done in the use of simple TAG; see Frank (1992) for a full discussion). However, the position in the projection of the overt direct object is not specified and it can move away from the verb, while still receiving $\theta$- and case-marking. Note that we have chosen to label all nodes in the projection of the verb with 'VP'; the functional information expressed by separate node labels in recent syntactic theories (IP, CP, AgrSP etc.) we express as features (not shown here for simplicity). We will not discuss the linguistic issues involved in using such sets as the formalism for representing linguistic competence (for details see Rambow and Lee (1994)).

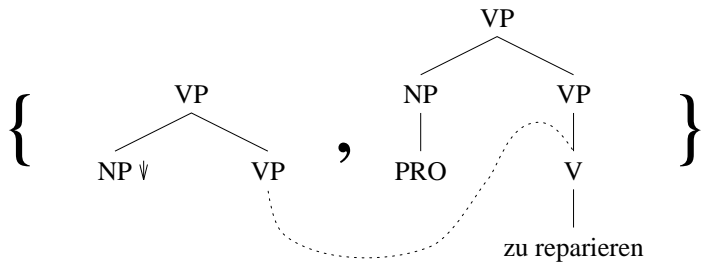

Figure 7: Multi-component tree set for *versprechen* 'to promise'

How can we derive sentences with long-distance scrambling? Consider (9e), repeated here for convenience:

(27) daß      [den Kühlschrank]$_i$      niemand      [[t$_i$ zu reparieren]    zu versuchen]    verpricht
     that     the refrigerator (ACC)     no-one (NOM) to repair                 to try           promises
     Comp$_1$ N$_3$                      N$_1$                                  V$_3$            V$_2$     V$_1$
     ...because no-one promises to repair the refrigerator

The accusative NP *den Kühlschrank* 'the refrigerator' has been scrambled out of the most deeply embedded clause into the matrix clause.[9] A complete grammar is given in Figure 8. There are two auxiliary trees for the matrix verb $V_1$ *versprechen* 'to promise' ($\beta_1$) and for the intermediate verb $V_2$ *versuchen* 'to try' ($\beta_2$). Finally, there is the tree set introduced earlier for the most deeply embedded verb $V_3$, *reparieren* 'to repair', containing auxiliary tree $\beta_{31}$ and initial tree $\alpha_{32}$. In the interest of readability, we use abbreviations for the terminal symbols, and we omit empty categories (PRO). These issues do not affect our discussion.

---

[8]In fact Linear Context-Free Rewriting Systems (LCFRS) are also not powerful enough. LCFRSs were introduced in (Weir, 1988) as a generalization of TAGs.

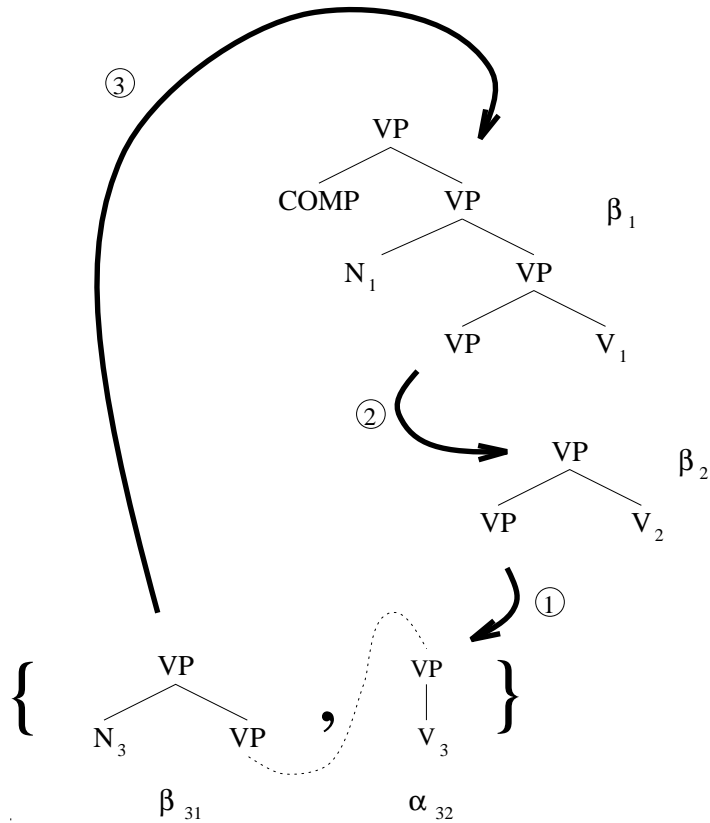

Figure 8: Grammar for German long-distance scrambling

The derivation is shown by the arrows in Figure 8. We start out by adjoining the intermediate clause into the verbal tree ($\alpha_{32}$) of the most deeply embedded clause at its root node, and then we adjoin the matrix clause into the intermediate clause at the root of the intermediate clause. Since we have not yet used the nominal argument tree from the most deeply embedded clause (tree $\beta_{31}$), the derivation is not yet complete. We choose to adjoin the most deeply embedded argument into the matrix clause, at the VP node between the complementizer and its nominal argument. This choice corresponds to long-distance scrambling. The resulting derived tree is shown in Figure 9.

Now let us turn to our processing model. Since the BEPDA is formally equivalent to TAG, and since TAG is formally inadequate for the long-distance phenomena we are interested in, we must extend our automaton model as well. We will do this by using an indexed version of the BEPDA, called the {}-BEPDA (Rambow, 1994). In the {}-BEPDA, every stack symbol in the pushdown store is associated with a set of indices. Intuitively, these indices represent trees that still need to be adjoined in order for the derivation to be successful. Since we are using auxiliary trees in sets to represent nominal arguments, we can think of these sets as unfulfilled (nominal) subcategorization requirements. We see that the notation is consistent with the quasi-categorial notation we adopted previously. If we do not allow stack symbols to pass subcategorization requirements to other stack symbols, we simply have a BEPDA which just recognizes Tree Adjoining Languages. However, if we allow symbols in the same stack of the push-down store to pass a subcategorization requirement to a stack symbol immediately above or below it, then we increase the power since we can now simulate the "detaching" of nominal arguments from their verbs. We will illustrate the functioning of the {}-BEPDA by showing how it performs on sentence (27).

---

[9]Of course, (27) is embedded in some other clause which we consistently omit in order to avoid the complications of the verb-second effect. Our use of the term 'matrix clause' to denote the topmost of the recursively embedded clauses is thus sloppy, but the intended meaning is clear.

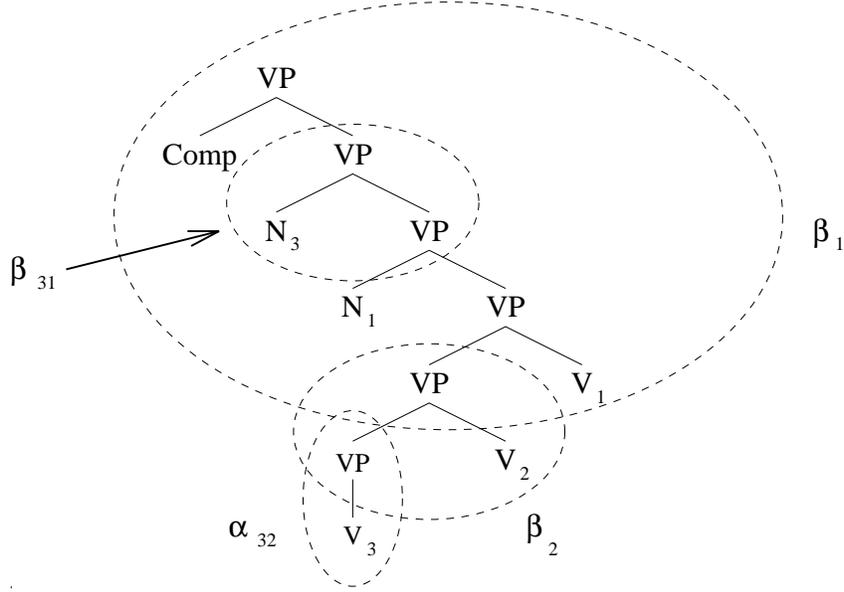

Figure 9: Derived tree for sentence (27)

(28)

| Step | Store | | | | | Input | New | Cum |
|---|---|---|---|---|---|---|---|---|
| 0 | | | | | | | 0 | 0 |
| 1 | [Comp$_1$ | | | | | Comp$_1$ | 1 | 1 |
| 2 | [Comp$_1$ | [N$_3$ | | | | N$_3$ | 2 | 3 |
| 3 | [Comp$_1$ | [N$_3$ | [N$_1$ | | | N$_1$ | 3 | 6 |
| 4 | [Comp$_1$ | [N$_3$ | [N$_1$ | [V$_3$\{N$_3$\} | | V$_3$ | 4 | 10 |
| 5a | [Comp$_1$ | [N$_3$ | [N$_1$ | [V$_3$\{N$_3$\} | [V$_2$\{\} | V$_2$ | 5 | 15 |
| 5b | [Comp$_1$ | [N$_3$ | [N$_1$ | [V$_3$\{N$_3$\} V$_2$\{\} | | | 5 | 20 |
| 6a | [Comp$_1$ | [N$_3$ | [N$_1$ | [V$_3$\{N$_3$\} V$_2$\{\} | [V$_1$\{N$_1$\} | V$_1$ | 6 | 26 |
| 6b | [Comp$_1$ | [N$_3$ | [N$_1$ | [V$_3$\{N$_3$\} V$_2$\{\} V$_1$\{N$_1$\} | | | 6 | 32 |
| 6c | [Comp$_1$ | [N$_3$ | [V$_3$\{N$_3$\} V$_2$\{\} V$_1$\{\} | | | | 6 | 38 |
| 6d | [Comp$_1$ | [N$_3$ | [V$_3$\{\} V$_2$\{N$_3$\} V$_1$\{\} | | | | 6 | 44 |
| 6e | [Comp$_1$ | [N$_3$ | [V$_3$\{\} V$_2$\{\} V$_1$\{N$_3$\} | | | | 6 | 50 |
| 6f | [Comp$_1$ | [V$_3$\{\} V$_2$\{\} V$_1$\{\} | | | | | 6 | 56 |
| 6g | [V$_3$\{\} V$_2$\{\} V$_1$\{\} | | | | | | 5 | 61 |
| 6h | [V$_3$\{\} V$_2$\{\} | | | | | | 3 | 64 |
| 6i | [V$_3$\{\} | | | | | | 2 | 66 |
| 6j | | | | | | | | 66 |

In steps 1 through 3, the complementizer and the two nouns are read in. In step 4, verb V$_3$ is read in, but no UNWRAP is possible, since V$_3$ is not next to its nominal argument. In steps 5a and 5b, verb V$_2$ is read in and UNWRAPped. In steps 6a, 6b, and 6c, the last verb V$_1$ is read in, and UNWRAPped around the stack of verbs and its nominal argument. At this point, no further reduction is possible without passing index symbols within a stack. This happens in steps 6d and 6e. When V$_1$ has "inherited" the subcategorization requirement of V$_3$, N$_3$ can be discharged. The fact that N$_3$ is UNWRAPped around V$_1$ corresponds to the fact that it has scrambled into the clause of V$_1$.

If we apply this method to the other sentences in (9) on page 4, we get the following results:

(29)

| No. | Sentence | Score | Judgment |
|---|---|---|---|
| (9b) | Comp$_1$ N$_1$ V$_1$ V$_2$ N$_3$ V$_3$ | 17 | ok |
| (9c) | Comp$_1$ N$_1$ V$_1$ N$_3$ V$_3$ V$_2$ | 24 | ok |
| (9a) | Comp$_1$ N$_1$ N$_3$ V$_3$ V$_2$ V$_1$ | 52 | (ok) |
| (9d) | Comp$_1$ N$_1$ N$_3$ V$_2$ V$_3$ V$_1$ | 58 | ? |
| (9e) | Comp$_1$ N$_3$ N$_1$ V$_3$ V$_2$ V$_1$ | 66 | ?? |

Sentence (9b) is fully extraposed: the score is low because material can be removed from the processor as soon as a clause is complete. In sentence (9c), the two most embedded clauses have been extraposed behind the matrix clause, but they have been left in a center-embedded construction. Thus the matrix clause can be removed from the automaton before any embedded clause is reached, but then the items must remain in the automaton until the whole sentence has been read in. Sentence (9a), which is prescriptively acceptable, is simply the fully center-embedded version; all lexical items must remain in the automaton until the entire sentence has been read in. This is also true for sentence (9d), but the score gets even worse, since $N_3$ has been long-distance scrambled out of the most deeply embedded clause (which has been extraposed) into the second clause. Finally, sentence (9e) is worst of all, since here there is long-distance scrambling over two clause boundaries. We see that the ordering of scores corresponds to the ordering by acceptability that we proposed earlier, and that, generally speaking, extraposition improves sentences, while long-distance scrambling degrades them.

Now let us address the second area that requires explanation, the apparent difference in processing load between long-distance topicalization and long-distance scrambling (over comparable distances). We repeat the contrasting sentences, first given as (10):

(30) a. Sentence with long-distance scrambling:

| ? | Der Meister | hat | den Kühlschrank | niemandem | zu reparieren | versprochen |
|---|---|---|---|---|---|---|
|  | the master | has | the refrigerator (ACC) | no-one (DAT) | to repair | promised |
|  | $N_{11}$ | $Aux_1$ | $N_2$ | $N_{12}$ | $V_2$ | $V_1$ |

The master has promised no-one to repair the refrigerator

b. Sentence with long-distance topicalization:

| Den Kühlschrank | hat | der Meister | niemandem | zu reparieren | versprochen |
|---|---|---|---|---|---|
| the refrigerator (ACC) | has | the master (NOM) | no-one (DAT) | to repair | promised |
| $N_2$ | $Aux_1$ | $N_{11}$ | $N_{12}$ | $V_2$ | $V_1$ |

The master has promised no-one to repair the refrigerator

Let us first consider the processing of the sentence with long-distance scrambling. The run of the automaton is similar to the one for sentence (27), except that we now have a full sentence with a matrix auxiliary in second position. We will assume it has been adjoined to the matrix verb and contributes features, in a manner similar to the complementizer in our previous example.

(31)

| Step | Store | | | | | | New | Cum |
|---|---|---|---|---|---|---|---|---|
| 1 | [$N_{11}$ | | | | | | 1 | 1 |
| 2 | [$N_{11}$ | [$Aux_1$ | | | | | 2 | 3 |
| 3 | [$N_{11}$ | [$Aux_1$ | [$N_2$ | | | | 3 | 6 |
| 4 | [$N_{11}$ | [$Aux_1$ | [$N_2$ | [$N_{12}$ | | | 4 | 10 |
| 5 | [$N_{11}$ | [$Aux_1$ | [$N_2$ | [$N_{12}$ | [$V_2\{N_2\}$ | | 5 | 15 |
| 6a | [$N_{11}$ | [$Aux_1$ | [$N_2$ | [$N_{12}$ | [$V_2\{N_2\}$ | [$V_1\{N_{11}, N_{12}\}$ | 6 | 21 |
| 6b | [$N_{11}$ | [$Aux_1$ | [$N_2$ | [$N_{12}$ | [$V_2\{N_2\}$ $V_1\{N_{11}, N_{12}\}$ | | 6 | 27 |
| 6c | [$N_{11}$ | [$Aux_1$ | [$N_2$ | [$V_2\{N_2\}$ $V_1\{N_{11}\}$ | | | 6 | 33 |
| 6d | [$N_{11}$ | [$Aux_1$ | [$N_2$ | [$V_2\{\}$ $V_1\{N_{11}, N_2\}$ | | | 6 | 39 |
| 6e | [$N_{11}$ | [$Aux_1$ | [$V_2\{\}$ $V_1\{N_{11}\}$ | | | | 6 | 45 |
| 6f | [$N_{11}$ | [$V_2\{\}$ $V_1\{N_{11}\}$ | | | | | 5 | 50 |
| 6g | [$V_2\{\}$ $V_1$ $\{\}$ | | | | | | 5 | 55 |
| 6h | [$V_2\{\}$ | | | | | | 2 | 57 |
| 6i | | | | | | | 0 | 57 |

Again, the long-distance scrambling is achieved by passing the subcategorization requirement from a verb to its governing verb (in step 6d), which represents the fact that $N_2$ has scrambled into the matrix clause. Now let us consider the run of the automaton in the topicalized case.

(32)

| Step | Store | | | | | | New | Cum |
|---|---|---|---|---|---|---|---|---|
| 1 | [$N_2$ | | | | | | 1 | 1 |
| 2 | [$N_2$ | [$Aux_1$ | | | | | 2 | 3 |
| 3 | [$N_2$ | [$Aux_1$ | [$N_{11}$ | | | | 3 | 6 |
| 4 | [$N_2$ | [$Aux_1$ | [$N_{11}$ | [$N_{12}$ | | | 4 | 10 |
| 5 | [$N_2$ | [$Aux_1$ | [$N_{11}$ | [$N_{12}$ | [$V_2\{N_2\}$ | | 5 | 15 |
| 6a | [$N_2$ | [$Aux_1$ | [$N_{11}$ | [$N_{12}$ | [$V_2\{N_2\}$ | [$V_1\{N_{11}, N_{12}\}$ | 6 | 21 |
| 6b | [$N_2$ | [$Aux_1$ | [$N_{11}$ | [$N_{12}$ | [$V_2\{N_2\}$ $V_1\{N_{11}, N_{12}\}$ | | 6 | 27 |
| 6c | [$N_2$ | [$Aux_1$ | [$N_{11}$ | [$V_2\{N_2\}$ $V_1\{N_{11}\}$ | | | 6 | 33 |
| 6d | [$N_2$ | [$Aux_1$ | [$V_2\{N_2\}$ $V_1\{\}$ | | | | 6 | 39 |
| 6e | [$N_2$ | [$V_2\{N_2\}$ $V_1\{\}$ | | | | | 5 | 44 |
| 6f | [$N_2$ | [$V_2\{N_2\}$ | | | | | 2 | 46 |
| 6g | $V_2\{\}$ | | | | | | 2 | 48 |
| 6h | | | | | | | 0 | 48 |

In the case of topicalization into sentence-initial position, we see that it is not necessary to pass subcategorization requirements among verbs. Instead, once the matrix clause has been removed (step 6e), the embedded verb is adjacent to its argument which can UNWRAP in the usual manner. This results in a lower score (48 as opposed to 57).

This difference in the behavior of the automata reflects the difference in the linguistic analysis in the competence grammar: while scrambling is achieved by adjoining NP arguments separately and in arbitrary order, topicalization is achieved by choosing a different elementary tree prior to the derivation. In topicalization, the long-distance effect is achieved by adjoining the matrix clause below the topicalized element, thus stretching it away from its verb. The difference in representation of scrambling and topicalization in the competence grammar is justified by the linguistic differences between the two word order variation types that we mentioned previously. (For example, anaphor binding behavior could be related to the fact that topicalization occurs within a single tree, while scrambling involves tree sets. We do not propose to work out the details of a TAG-based binding theory here.) Once again, we see that the independently motivated competence theory leads to automata models that make highly plausible predictions.

## CONCLUSION

We have presented a model of human syntactic processing that makes plausible predictions for a range of word-order variation phenomena in German (and Dutch). Our model of the human syntactic processor is directly linked to a TAG-based model of human syntactic competence. This direct link gives our model two major characteristics that differentiate it from other models:

- The processor is not concerned directly with phrase-structure trees, but with relations between lexical items.

- The processor is defined in terms of a set of formally defined operations, which may at first appear arbitrary.

We will briefly discuss these two points in turn, and finish with a brief note on syntactic ambiguity.

The parser simulates a derivation in the formalism of the competence theory, TAG. TAG is a tree-rewriting system, and therefore derivations in a TAG are not recorded by a phrase-structure tree (as is the case for CFG), but by the so-called *derivation tree*, which is a tree that represents adjunctions and substitutions performed during the derivation. Each node in the derivation tree corresponds to one elementary tree, and a dominance relation represents adjunction (or substitution) of the tree represented by the daughter node into the mother. Since the representation of competence exploits the lexicalizability of TAG, each tree in our competence grammar is associated with one lexical item. This means that the parser is in fact establishing direct dependencies between lexical items (heads). However, our approach does not build a phrase-structure tree (though one can be derived from its actions, just as one can be derived form the derivation tree). While explanatory approaches such as Minimal Attachment are less appealing in our model (since no phrase-structure tree is explicitly represented during the parse), licensing properties of lexical items can be represented in straightforward manner. In this respect, our approach is close to licensing-based or head-driven approaches (Abney, 1986; Pritchett, 1991). Lexical licensing relations, in particular $\theta$-role assignment, also play a crucial role in approaches that are not "head-driven" (Gibson, 1991; Inoue and Fodor, 1993). We suspect that such conditions will find a straightforward representation in our processing model.

Furthermore, the lexicon-oriented processing model allows for an elegant integration of lexical co-occurrence effects, which, it is generally believed, play a crucial role in parsing.

The second characteristic of our model that we would like to discuss is its very precise definition, which may seem somewhat arbitrary at first: why does the push-down store contain stacks of stack symbols, and why may each stack symbol be associated with an index set? The justification for this machinery comes from a careful study of the requirements of competence syntax. It is known that (case-marked) cross-serial dependencies are not context-free (Shieber, 1985) – therefore, the representation of competence syntax cannot be based on a transformation-free CFG, nor can the parser be, say, a simple PDA. The representation of competence must therefore either include transformations, in which case we give up formal constraints and any hope that a formal analysis can guide us in modeling the parser, or we can look for other (more powerful yet still constrained) formalisms for the expression of competence. We will not argue for our choice (TAG and its extension, VMC-TAG-DL) here (the reader is referred to (Joshi, 1985) and (Becker, Joshi, and Rambow, 1991; Rambow, 1994), respectively), but we claim that the complexity of the machinery of the processing model is justified by the details of the competence model. If the processing model makes empirically interesting predictions, then the complexity of its operations and data structures should not be held against it (on the basis of scientific parsimony) since they are independently motivated.

Finally, we need to address the issue of ambiguity. The model, as presented in (Joshi, 1990) and extended and modified here, does not address the issue of the resolution of syntactic ambiguity. (The reader will have observed that in all of our examples, the syntactic structure is in fact unambiguous – partly due to the case-marking. Furthermore, in all cases, the processing difficulties are not of the garden-path variety, since they persist even when the reader is primed for the syntactic structure.) *A priori*, it seems that our model can be integrated into a variety of ambiguity-resolution models, including parallelism, limited parallelism, deterministic with lookahead, and serial with (limited) backtracking. In further work, we intend to investigate whether the particular features of our model favor one or the other of these approaches.

# Notes